# Toward General Analysis of Recursive Probability Models

**Daniel Pless and George Luger**
Computer Science Department
University of New Mexico
Albuquerque NM 87131
{dpless, luger}@cs.unm.edu

## Abstract

There is increasing interest within the research community in the design and use of recursive probability models. There remains concern about computational complexity costs and the fact that computing exact solutions can be intractable for many nonrecursive models. Although inference is undecidable in the general case for recursive problems, several research groups are actively developing computational techniques for recursive stochastic languages. We have developed an extension to the traditional λ-calculus as a framework for families of Turing complete stochastic languages. We have also developed a class of exact inference algorithms based on the traditional reductions of the λ-calculus. We further propose that using the deBruijn notation (a λ-calculus notation with nameless dummies) supports effective caching in such systems, as the reuse of partial solutions is an essential component of efficient computation. Finally, our extension to the λ-calculus offers a foundation and general theory for the construction of recursive stochastic modeling languages as well as promise for effective caching and efficient approximation algorithms for inference.

## 1   INTRODUCTION

The limitations of flat Bayesian Networks (BNs) using simple random variables have been widely noted by researchers [Xiang et al., 1993; Laskey and Mahoney, 1997]. These limitations have motivated a variety of recent research projects in hierarchical and composable Bayesian models [Koller and Pfeffer, 1997; Koller and Pfeffer, 1998; Laskey and Mahoney, 1997; Pfeffer et al., 1999; Xiang et al., 2000]. Most of these new Bayesian modeling formalisms support model decomposition, often based on an object-oriented approach. Although these approaches provide more expressive and/or succinct representational frameworks, few of these change the class of models that can be represented.

Recent research has addressed this issue. One example is the functional stochastic modeling language proposed by Koller et al. [1997]. Their language is Turing complete, allowing the representation of a much broader class of models. Pless et al. [2000] extends and refines this proposed framework to one which is more object-oriented and which allows hierarchical encapsulation of models. Both languages provide the ability to use functions to represent general stochastic relationships. They both also use lazy evaluation to allow computation over potentially infinite distributions. Pfeffer [2000] and Pfeffer and Koller [2000] have also proposed a Turing complete framework based on approximate inference.

Another approach to the representation problem for stochastic inference is the extension of the usual propositional nodes for Bayesian inference to the more general language of first-order logic. Kersting and De Raedt [2000] associated first-order rules with uncertainty parameters as the basis for creating Bayesian networks as well as more complex models. Poole [1993] gives an earlier approach which develops an approximate algorithm for another Turing complete probabilistic logic language.

These approaches all have in common the development of recursive models that bring together inference in Bayesian Networks with more complex models such as stochastic context free grammars. The result aims at allowing the construction and inference in novel Bayesian models. All of these methods depend on caching of partial results for efficiency purposes, just as efficient inference in Bayesian Networks requires the storage of intermediate values.

In this paper we offer an extension to the traditional λ-calculus that provides a foundation for building Turing complete stochastic modeling languages. We have also developed a class of exact stochastic inference algorithms based on the traditional reductions in the λ-calculus. We further propose the use of deBruijn [1972] notation to support effective caching mechanisms for efficient computation. As noted above, caching offers an important technique for support of efficient inference in stochastic networks.

The problem with using the λ-calculus directly is that it is quite natural to develop two or more expressions that are equivalent, only differing in the choice of bound variable names. Furthermore, variable substitution is complicated by the requirement of variable binding and capture, which often makes substitutions quite expensive.

deBruijn notation addresses both of these issues by replacing arbitrary variable names with explicitly specified positive integers, which addresses the naming problem. It simultaneously renders the variable capture problem easy, by eliminating the arbitrary variable names that can be accidentally bound, and thus makes the substitution problem less computationally expensive.

As a final note, other recent research has viewed stochastic modeling in terms of stochastic functions [Pearl, 2000; Koller, and Pfeffer, 1997]. For example, Pearl's [2000] recent book constructs a formalism for "causality" in terms of stochastic functions. We have expanded these ideas based on an extension of the λ-calculus, in which the stochastic functions themselves become first class objects, to offer a formal structure for such modeling.

## 2    THE EXTENDED λ-CALCULUS FORMALISM

We now present a formal grammar reflecting our extension of the λ-calculus to describe stochastic distributions. The goal of this effort is to propose an extended form that also supports an inference algorithm as a set of standard transformations and reductions of λ-calculus forms. Thus, inference in our modeling language is equivalent to finding normal forms in the λ-calculus. We also modify our language through the use of deBruijn notation [deBruijn, 1972]. This notation replaces arbitrarily chosen variable names with uniquely determined positive integers. As a result all expressions that are α-equivalent in standard notation are identical under deBruijn notation. This is very useful in both constructing distributions as well as in caching partial results.

### 2.1    Syntax

We next present a pseudo-BNF grammar to describe our stochastic extension of the traditional λ-calculus:

```
<expr> ::=
          <var> | <λ> | <application> | <distribution>
<var> ::= <integer>
<λ> ::= (λ <expr>)
<application> ::= (<expr>₁ <expr>₂)
<distribution> ::= ∑ᵢ <expr>ᵢ: <p>ᵢ
p ∈ (0, 1]
```

Thus, our stochastic λ-calculus contains the same elements as standard λ-calculus: variables, λ-abstractions, and function applications. In addition, in our stochastic λ-calculus, it is legal to have an expression which is itself a distribution of expressions.

One difficulty mentioned earlier with standard λ-calculus is that there is an unfortunate representational indeterminacy in the arbitrary choice of bound variable names. Thus two completely equivalent (under α-rule) expressions can have different forms. This presents a problem as we need to be able to combine probabilities of the same expression occurring within a distribution. Therefore we use deBruijn notation to give each expression a canonical form. This canonical form also makes $O(1)$ caching of the evaluation of expressions possible.

Another advantage of deBruijn notation is that it simplifies substitution (as discussed in section 3) and allows for the reuse of entire sub-expressions, which allows faster substitution when performing λ reductions.

It should be noted that deBruijn [1972] proposed the notation as an improvement for machine manipulation of expressions and not for human use. Our purpose in developing the stochastic λ-calculus is to provide the expressiveness of a higher order representation and an effective framework for inference. We are not aiming to develop a user-friendly language in this paper. In actual model development, one might use a high level language similar to the other languages discussed in the introduction, and then compile that language to our stochastic λ-calculus.

When using deBruijn notation, we denote a variable by a positive integer. This number indicates how many λs one must go out to find the one λ to which that variable is bound. We denote a λ-abstraction with the form (λ e) where e is some legal expression. For example (λ 1) represents the identity function. In λ-calculus, boolean values are often represented by functions that take two arguments. true returns the first one, and false the second. In this notation true becomes (λ (λ 2)) and false is (λ (λ 1)), or in an abbreviated form (λλ 2) and (λλ 1) respectively.

For a further example we use deBruijn notation to describe the S operator from combinatory logic. The S operator may be described by the rule Sxyz = (xz)(yz) which is equivalent to the standard λ term (λxλyλz.(xz)(yz)). In deBruijn notation, this becomes (λλλ (3 1)(2 1)).

Function application is as one might expect: We have $(e_1\ e_2)$, where $e_1$ is an expression whose value will be applied as a function call on $e_2$, where $e_2$ must also be a valid expression. We describe distributions as a set of expressions annotated with probabilities. An example would be a distribution that is 60% true and 40% false. Using the representation for boolean values given above, the resulting expression would be: {(λλ 2): 0.6, (λλ 1): 0.4}. Note that we use a summation notation in our BNF specification. The set notation is convenient for denoting a particular distribution, while the summation notation is better for expressing general rules and algorithms.

## 2.2   SEMANTICS

We next develop a specification for the semantics of our language. For expressions that do not contain distributions, the semantics (like the syntax) of the language is the same as that of the normal λ-calculus. We have extended this semantics to handle distributions.

A distribution may be thought of as a variable whose value will be determined randomly. It can take on the value of any element of its set with a probability given by the annotation for that element. For example, if T denotes true as represented above, and F represents false, the distribution {T: 0.6, F: 0.4} represents the distribution over true and false with probability 0.6 and 0.4 respectively.

A distribution applied to an expression is viewed as equivalent to the distribution of each element of the distribution applied to the expression, weighted by the annotated probability. An expression applied to a distribution is likewise the distribution of the expression applied to each element of the distribution annotated by the corresponding probability. Note that in both these situations, when such a distribution is formed it may be necessary to combine syntactically identical terms by adding the annotated probabilities.

Under our grammar it is possible to construct distributions that contain other distributions directly within them. For example {T: 0.5, {T: 0.6, F: 0.4}: 0.5} is the same as {T: 0.8, F: 0.2}. One could explicitly define a reduction (see below) to handle this case, but in this paper we assume that the construction of distributions automatically involves a similar flattening just as it involves the combination of probabilities of syntactically identical terms.

In other situations an application of a function to an expression follows the standard substitution rules for the λ-calculus with one exception: The substitution cannot be applied to a general expression unless it is known that the expression is not reducible to a distribution with more than one term. For example, an expression of the form ((λ $e_1$) (λ $e_2$)) can always be reduced to an equivalent expression by substituting $e_2$ into $e_1$ because (λ $e_2$) is not reducible. We describe this situation formally with our presentation of the reductions in the next section on stochastic inference.

There is an important implication of the above semantics. Every application of a function whose body includes a distribution causes an independent sampling of that distribution. There is no correlation between these samples. On the other hand, a function applied to a distribution induces a complete correlation between instances of the bound variables in the body of the function.

For example, using the symbols T and F as described earlier, we produce two similar expressions. The first version, (λ 1 F 1){T: 0.6, F: 0.4}, demonstrates the induced correlations. This expression is equivalent to F (false). This expression is always false because the two

1's in the expression are completely correlated (see the discussion of the inference reductions below for a more formal demonstration). Now to construct the second version, let G = (λ {T: 0.6, F: 0.4}). Thus G applied to anything produces the distribution {T: 0.6, F: 0.4}. So the second version ((G T) F (G T)) looks similar to the first one in that they both look equivalent to ({T: 0.6, F: 0.4} F {T: 0.6, F: 0.4}). The second version is equivalent because separate calls to the same function produce independent distributions. The first is not equivalent because of the induced correlation.

Finally, it should be noted that we can express Bayesian Networks and many other more complex stochastic models, including Hidden Markov Models, with our language. Using the Y operator of combinatory logic [Hindley and Seldin, 1989], any recursive function can be represented in standard as well as stochastic λ-calculus. This operator has the property that Yf = f(Yf) in the standard λ-calculus. In our extended formalism, the Y operator is no longer a fixed-point operator for all expressions, but can be used to construct recursive functions. Thus the language is Turing complete, and can represent everything that other Turing complete languages can. For illustration, we next show how to represent the traditional Bayesian Network in our stochastic λ-calculus.

## 2.3   AN EXAMPLE: REPRESENTING BAYESIAN NETWORKS (BNs)

To express a BN, we first construct a basic expression for each variable in the network. These expressions must then be combined to form an expression for a query. At first we just show the process for a query with no evidence. The technique for adding evidence will be shown later. A basic expression for a variable is simply a stochastic function of its parents.

To form an expression for the query, one must form each variable in turn by passing in the distribution for its parents as arguments. When a variable has more than one child, an abstraction must be formed to bind its value to be passed to each child separately.

Our example BN has three Boolean variables: A, B, and C. Assume A is true with probability of 0.5. If A is true, then B is always true, otherwise B is true with probability of 0.2. Finally, C is true when either A or B is true. Any conditional probability table can be expressed in this way, but the structured ones given in this example yield more terse expressions. The basic expressions (represented with both standard and deBruijn notation) are shown below:

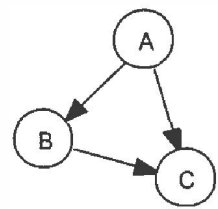

$A = \{T: 0.5, F: 0.5\}$
$B = (\lambda A.(A\ T\ \{T: 0.2, F: 0.8\}))$
  $= (\lambda\ 1\ T\ \{T: 0.2, F: 0.8\})$
$C = (\lambda A \lambda B.(A\ T\ B)) = (\lambda \lambda\ 2\ T\ 1)$

The complete expression for the probability distribution for C is then $((\lambda\ C\ 1\ (B\ 1))\ A)$. One can use this to express the conditional probability distribution that A is true given that C is true: $((\lambda\ (C\ 1\ (B\ 1))\ 1\ N)\ A)$ where N is an arbitrary term (not equivalent to T or F) that denotes the case that is conditioned away. To infer this probability distribution, one can use the reductions (defined below) to get to a normal form. This will be a distribution over T, F, and N, with the subsequent marginalizing away of N.

In general, to express evidence, one can create a new node in the BN with three states. One state is that the evidence is false, the second is the evidence and the variable of interest are true, and the third represents the evidence true and the variable of interest is false. One can then get the distribution for the variable of interest by marginalizing away the state representing the evidence being false. The extension to non boolean variables of interest is straightforward.

Of course, a language with functions as first class objects can express more than Bayesian Networks. It is capable of expressing the same set of stochastic models as the earlier Turing complete modeling languages as proposed [Koller et al., 1997; Pless et al., 2000; Pfeffer, 2000; Pfeffer and Koller, 2000]. Any of those languages could be implemented as a layer on top of our stochastic $\lambda$-calculus. In Pless et al. [2000] the modeling language is presented in terms of an outer language for the user which is then transformed into an inner language appropriate for inference. Our stochastic $\lambda$-calculus could also be used as a compiled form of a more user friendly outer language.

In summary, we have created a Turing complete specification for representing stochastic reasoning. We have proposed an extension to the standard $\lambda$-calculus under deBruijn notation to give an effective form for inference (discussed in the next section). Our specification can be used to de-couple the design of high level stochastic languages from the development of efficient inference schemes.

## 3   STOCHASTIC INFERENCE THROUGH $\lambda$ REDUCTIONS

We next describe exact stochastic inference through the traditional methodology of the $\lambda$-calculus, a set of $\lambda$ reductions. In addition to the $\beta$ and $\eta$ reductions, we also define a new form: $\gamma$ reductions.

$\beta$:  $((\lambda\ e_1)\ e_2) \rightarrow$ substitute$(e_1, e_2)$
$\gamma_L$: $((\Sigma_i\ f_i: p_i)\ e) \rightarrow \Sigma_i\ (f_i\ e): p_i$
$\gamma_R$: $(f\ \Sigma_i\ e_i: p_i) \rightarrow \Sigma_i\ (f\ e_i): p_i$
$\eta$:  $(\lambda\ (e\ 1)) \rightarrow e$

We have defined $\beta$ reductions in a fashion similar to standard $\lambda$-calculus. Since we are using deBruijn notation, $\alpha$ transformations become unnecessary (as there are no arbitrary dummy variable names). $\beta$ and $\eta$ reductions are similar to their conventional counterparts (see deBruijn [1972] for restrictions on when they may be applied). In our case the difference is that $\beta$ reductions are more restricted in that expressions that are reducible to distributions cannot be substituted. Similarly $\eta$ reductions are restricted to those expressions $(\lambda\ (e\ 1))$ where e cannot be reduced to a distribution. In addition to those two standard reductions we define two additional reductions that we term $\gamma_L$ and $\gamma_R$. The $\gamma$ reductions are based on the fact that function application and distributions distribute.

One important advantage of using deBruijn notation is the ability to reuse expressions when performing substitutions. We next present a simple algorithm for substitutions when $e_2$ is a closed expression:

level(expr) = case expr
    var $\rightarrow$ expr
    $(\lambda\ e) \rightarrow$ max(level(e) $-$ 1, 0)
    $(e_1\ e_2) \rightarrow$ max(level$(e_1)$, level$(e_2)$)
    $\Sigma_i\ e_i: p_i \rightarrow$ max$_i$(level$(e_i)$)

substitute$((\lambda\ e_1),\ e_2)$ = substitute$(e_1, e_2, 1)$

substitute(expr, a, L) = if level(expr) < L then expr
    else case expr
        var $\rightarrow$ a
        $(\lambda\ e) \rightarrow$ $(\lambda$substitute(e, a, L+1))
        $(e_1\ e_2) \rightarrow$ (substitute$(e_1,$ a, L)
                         substitute$(e_2,$ a, L))
        $\Sigma_i\ e_i: p_i \rightarrow \Sigma_i$ substitute$(e_i,$ a, L): p_i

This algorithm is designed to maximize the reuse of sub-expressions. When a new expression is built, a non-negative integer value, called the level, is associated with it. This value is the maximum number of $\lambda$s that have to surround the expression for it to be closed. The level is defined recursively, and is calculated directly from the sub-expressions from which the newly created expression is derived.

For a variable, its level is the number denoting the variable. For a $\lambda$-abstraction, the level is derived from the level of the expression in the body of the abstraction, but reduced by one (to a minimum of zero) due to the $\lambda$. For forms that combine expressions (applications and distributions) the level is the maximum level of the sub-expressions being combined. The level function reflects this recursive construction.

The level value (whose construction doesn't increase the asymptotic time for building expressions) is valuable for avoiding unnecessary substitutions. The substitute function defines how to substitute an expression $e_2$ into a $\lambda$-abstraction $(\lambda\ e_1)$. This results in a call to the three parameter recursive version of substitute. The first parameter is the expression being substituted into, the second is the expression being substituted, and the last is the variable value that has to be replaced by the second argument.

If the variable to be substituted is greater than the level of the expression, then there cannot be any substitutions needed as the variable number is greater than the number of λs required to close the expression. In this case the expression can be directly returned (and reused). Otherwise, if the expression to be substituted is a variable, under the assumption that the original expression was closed, it must be the variable to be replaced. Thus, the substituting expression (second argument) can be returned (and reused). For λ-abstractions, the substitution is performed on the body of the abstraction, but substituting for a variable one larger in value. The result is then placed back into a λ-abstraction. Finally, for the combining forms, the substitution is performed on all of the sub-expressions and then recombined.

As noted earlier, we have defined two additional reductions that we call $\gamma_L$ and $\gamma_R$. The $\gamma_R$ reduction is essential for reducing applications where the β reduction cannot be applied. Continuing the example introduced earlier in the paper:

$$(\lambda\ 1\ F\ 1)\{T:\ 0.6,\ F:\ 0.4\} \xrightarrow{\gamma_R}$$
$$\{((\lambda\ 1\ F\ 1)\ T):\ 0.6,\ ((\lambda\ 1\ F\ 1)\ F):\ 0.4\}$$

Now since both T and F do not contain distributions, β reductions can be applied:

$$\{((\lambda\ 1\ F\ 1)\ T):\ 0.6,\ ((\lambda\ 1\ F\ 1)\ F):\ 0.4)\} \xrightarrow{\beta}$$
$$\{(T\ F\ T):\ 0.6,\ ((\lambda\ 1\ F\ 1)\ F):\ 0.4\}$$

$$\{(T\ F\ T):\ 0.6,\ ((\lambda\ 1\ F\ 1)\ F):\ 0.4\} \xrightarrow{\beta}$$
$$\{(T\ F\ T):\ 0.6,\ (F\ F\ F):\ 0.4\}$$

And now, using the definitions of T and F it is easy to see that (T F T) and (F F F) both are reducible to F.

# 4 INFERENCE

The task of inference in our stochastic λ-calculus is the same as the problem of finding a normal form for an expression. In standard λ-calculus, a normal form is a term to which no β reduction can be applied. In the stochastic version, this must be modified to be any term to which no β or γ reduction can be applied. It is a relatively simple task to extend the Church-Rosser theorem [Hindley and Seldin, 1986; deBruijn, 1972] to show that this normal form, when it exists for a given expression, is unique. Thus one can construct inference algorithms to operate in a manner similar to doing evaluation in a λ-calculus system. Just as it is possible to produce complete function evaluation algorithms in standard λ-calculus, the stochastic λ-calculus admits complete inference schemes.

## 4.1    A SIMPLE INFERENCE ALGORITHM

We next show a simple algorithm for doing such evaluation. This algorithm doesn't reduce to a normal form, rather to the equivalent of a weak head normal form [Reade, 1989].

```
peval(expr) = case expr
    (λ e) → expr
    (e₁ e₂) → papply(peval(e₁), e₂)
    Σᵢ eᵢ: pᵢ → Σᵢ peval(eᵢ): pᵢ

papply(f, a) = case f
    Σᵢ fᵢ: pᵢ → Σᵢ papply(fᵢ, a)::pᵢ
    (λ fₑ) → case a
        (λ e) → peval(substitute(f, a))
        (e₁ e₂) → papply(f, peval(a))
        Σᵢ eᵢ: pᵢ → Σᵢ papply(f, eᵢ): pᵢ
```

peval and papply are the extended version of eval and apply from languages such as LISP. peval implements left outermost first evaluation for function applications ((e₁ e₂)). For λ-abstractions, (λ e), no further evaluation is needed (it would be if one wanted a true normal form). For distributions, it evaluates each term in the set and then performs a weighted sum.

papply uses a $\gamma_L$ reduction when a distribution is being applied to some operand. When a λ-abstraction is being applied, its behavior depends on the operand. When the operand is an abstraction, it applies a β reduction. If the operand is an application, it uses eager evaluation (evaluating the operand). When the operand is a distribution, it applies a $\gamma_R$ reduction.

## 4.2    EFFICIENCY ISSUES

We have presented this simple, but not optimal, algorithm for purposes of clarity. One key problem is that it uses lazy evaluation only when the operand is a λ-abstraction. One would like to use lazy evaluation as much as possible. An obvious improvement would be to check to see if the bound variable in an operator is used at least one time. If it is not used then it doesn't matter whether the expression evaluates to a distribution or not, lazy evaluation can be applied.

Another potential improvement is to expand the set of cases in which it is determined that the operand cannot be reduced to a distribution. To make this determination in all cases is as hard as evaluating the operand, which is exactly what one tries to avoid through lazy evaluation. However, some cases may be easy to detect. For example, an expression that doesn't contain any distributions in its parse tree clearly will not evaluate to a distribution. One approach might be to use a typed λ-calculus to identify whether or not an expression could be reduced to a distribution.

Finally, we may tailor the algorithm using the reductions in different orders for particular application domains. The algorithm we presented doesn't utilize the η reduction, which may help in some cases. Also identifying more cases when β reductions can be applied may allow for more efficient algorithms in specific applications.

We propose that employing the simple algorithm with the suggested improvements (both shown above) will essentially replicate the variable elimination algorithm for inference on BNs. The order for variable elimination is implicitly defined by the way that the BN is translated into a λ-expression. The use of λ-expressions to form conditional probability tables also allows the algorithm to exploit context specific independence [Boutilier et al, 1996].

### 4.3 CACHING

Efficient computational inference in probabilistic systems generally involves the saving and reuse of partial and intermediate results [Koller et al., 1997]. Algorithms for BBNs as well as for HMMs and other stochastic problems are often based on some form of dynamic programming [Dechter, 1996, Koller et al., 1997]. Using deBruijn notation makes caching expressions easy. Without the ambiguity that arises from the arbitrary choice of variable names (α-equivalence), one needs only to find exact matches for expressions.

Because it is possible in λ-calculus to use the Y (fixed-point) operator from combinatory logic to represent recursion, there are no circular structures that need to be cached. Thus, only trees need be represented for caching of purely deterministic expressions. To cache distributions (which is necessary for non-deterministic caching) one needs to be able to cache and retrieve sets.

One way to accomplish this is to use the hashing method of Wegman and Carter [1981]. They propose a probabilistic method for producing hash values (fingerprints) for sets of integers. Their method is to associate each integer appearing in some set with a random bit string of fixed length. The bit strings for a particular set of integers are combined using the exclusive-or operation. Two sets are considered to be the same if the fingerprints for the two are the same. There is a probability that a false match can be found this way, which can be made arbitrarily low by increasing the string length.

This method can be used for caching the sets of weighted expressions (distributions) by associating a random bit string with each probability-expression pair that exists in some distribution. One can use a similar assignment of random strings to the integers representing variables. Also such a string can be assigned to fingerprint the λ in λ-abstractions. One can assign such fingerprints to unique pairs that occur in building up expressions. In this way one can form a hash function which (given the values of the sub-expression from which the expression is formed) can be computed without increasing the asymptotic time for expression construction.

## 5. APPROXIMATION

One of the strengths of viewing stochastic inference in terms of the λ-calculus with reductions is that it allows analysis of different parts of the expression to be handled differently. One way that can occur is to use different reduction orders on different parts of the expression. A powerful approach is to use approximations on different parts of the expression.

One may choose at some point in the evaluation to replace an expression with a reasonable distribution over the possible values the expression could potentially evaluate. Doing this at a fixed recursion level in the algorithm suggested above essentially gives the approximation suggested by Pfeffer and Koller [2000]. Other approximations include removing low probability elements from a distribution prior to performing a γ reduction. This is analogous to the approximation for Bayesian Networks proposed by Jensen and Anderson [1990]. Furthermore, a portion of the expression may be sampled with a Monte-Carlo algorithm.

Finally one can perform an improper β reduction when it is not allowed under stochastic λ-calculus: namely an application where the argument is reducible to a distribution and the function uses the argument more than once in its body. This last approximation corresponds to making an independence assumption that isn't directly implied by the form of the expression. That is, it assumes that the different instances of the argument in the body of the function are independent. The stochastic λ-calculus provides a framework for mixing and combining all of these different forms of approximation.

## 6. CONCLUSIONS AND FUTURE WORK

We have presented a formal framework for recursive modeling languages. It is important to maintain the distinction between our modeling language and a traditional programming language. Our language is designed to construct and satisfy queries on stochastic models, not to build programs. Some of our design decisions follow from this fact. We have made function application on a distribution result in essentially sampling from that distribution. If one wants to pass a distribution to a function as an abstract object, rather than sample from it, one must wrap the distribution in a λ-abstraction. We believe that the use of distributions without sampling will be rare unless the distribution is parameterized, in which case a function is needed anyway.

The result of the above decision is that the concept of abstract equality of expressions in our formalism is not the same as in standard λ-calculus. In standard λ-calculus, equality between expressions can be defined in terms of equality of behavior when the expressions are applied to arguments. In our λ-calculus, it is possible for two expressions to behave the same way when applied to any argument, but to differ in behavior when used as an argument to some other expression.

There are a number of paths that would be interesting to follow. It would be useful to analyze the efficiency of various algorithms on standard problems, such as polytrees [Pearl, 1988], where the efficiency of the optimal algorithm is known. This may point to optimal reduction orderings and other improvements to inference. We are also looking at constructing formal models of the semantics of the language. Finally, we are considering the implications of moving from the pure λ-calculus presented here to an applicative λ-calculus. The results of that representational change, along with type inference mechanisms, may be important for further development in the theory of recursive stochastic modeling languages.

## Acknowledgements

This work was supported by NSF Grant 115-9800929. We would also like to thank Carl Stern and Barak Pearlmutter for many important discussions on our approach.